\documentclass[final,5p,times,twocolumn]{elsarticle}

\usepackage{amssymb}
\usepackage{amsmath}
\usepackage{subfigure}
\usepackage{multirow}

\begin{document}

\begin{frontmatter}



\title{Object detection-based inspection of power line insulators: Incipient fault detection in the low data-regime}

\author[ETHZ]{Laya Das}
\author[ETHZ]{Mohammad Hossein Saadat}
\author[ETHZ]{Blazhe Gjorgiev}
\author[SG]{Etienne Auger}
\author[ETHZ]{Giovanni Sansavini\corref{*}}

\cortext[*]{Corresponding author, email: sansavig@ethz.ch}
\affiliation[ETHZ]{organization={Reliability and Risk Engineering},
            addressline={Department of Mechanical and Process Engineering}, 
            city={Zurich},
            postcode={8092}, 
            country={Switzerland}}

\affiliation[SG]{organization={Swissgrid AG},
            city={Aarau},
            postcode={5001}, 
            country={Switzerland}}

\begin{abstract}
Deep learning-based object detection is a powerful approach for detecting faulty insulators in power lines. This involves training an object detection model from scratch, or fine tuning a model that is pre-trained on benchmark computer vision datasets. This approach works well with a large number of insulator images, but can result in unreliable models in the low data regime. The current literature mainly focuses on detecting the presence or absence of insulator caps, which is a relatively easy detection task, and does not consider detection of finer faults such as flashed and broken disks. In this article, we formulate three object detection tasks for insulator and asset inspection from aerial images, focusing on incipient faults in disks. We curate a large reference dataset of insulator images that can be used to learn robust features for detecting healthy and faulty insulators. We study the advantage of using this dataset in the low target data regime by pre-training on the reference dataset followed by fine-tuning on the target dataset. The results suggest that object detection models can be used to detect faults in insulators at a much incipient stage, and that transfer learning adds value depending on the type of object detection model. We identify key factors that dictate performance in the low data-regime and outline potential approaches to improve the state-of-the-art.
\end{abstract}




\end{frontmatter}


\section{Introduction}
Power transmission insulators are an important component of the transmission system and play a key role in ensuring reliable supply of power. They act as a bridge between transmission lines and transmission towers allowing for physical support, while preventing leakage of current through the tower~\cite{werneck2014detection}. Much like other components of the transmission grid, they are subject to day-to-day variations in operation caused by fluctuating demand and generation, as well as to extreme environmental conditions such as lightning strikes, wind storms and rain. These factors, in addition to deposition of dust, rising daily temperatures and rapidly changing weather conditions deteriorate the quality of insulation, thereby allowing current to leak~\cite{li2010,ramirez2012}. The leakage current can worsen over time and ultimately result in a protection device to activate and disconnect the line~\cite{lisun2009}. Regular inspection and maintenance of insulators is therefore important to ensure reliable operation of the system.

There is an increasing interest in exploiting powerful deep learning models to perform automated monitoring of the health of insulators. One class of these studies relies on patterns manifested in measurements of current (or derivatives thereof) from multiple towers, and makes use of machine learning and deep learning methods for pattern recognition \cite{gjorgiev2023LC}. The second class of studies, and the topic considered in this article, involves collection of aerial images of insulators and uses deep learning models for detection and classification of insulators. Such an automated process reduces the reliance on human experts to inspect insulators and the associated costs and human errors. This also increases the safety of the data acquisition process, which involves controlling a drone from a distance instead of physically climbing the tower to collect the images.

\subsection{Related Work}
Several articles in the literature have developed machine learning and deep learning-based solutions for assessing the state of insulators from aerial images. The authors in \cite{junfeng2017novel} use saliency maps for image segmentation with $200$ images through static local and global feature extraction. A weighted combination of the local and global saliency maps is used to extract insulators in an image. A Faster RCNN object detection model is applied to detect three different types of insulators in \cite{liu2018insulator}. A dataset with $1000$ images for each type of insulator is used to train the model, which is then tested on $500$ images of each type of insulator. The authors also carry out fault detection with missing caps of insulators with the model, which is trained with $80$ images and tested on $40$ images, yielding $92\%$ precision. The Single-Shot multi-box Detector (SSD) is utilized in \cite{jiang2019insulator} to detect healthy and faulty insulators. The model is trained with $385$ images and achieves a precision of $92.48\%$ on a test dataset of $100$ images. The authors in \cite{han2019method} employ a modified object detector inspired from the highly successful You Only Look Once version 2 (YOLOv2) model to detect missing disks in insulators. The authors compile a dataset of $4031$ images and use modified YOLO models to achieve a detection precision of $94.2\%$ for insulators with single missing disk and $98.3\%$ for insulators with multiple missing disks. Similar approaches that make modifications to the model architecture to improve the performance of detection are proposed in the literature \cite{liu2021insulator,liu2021improved,xia2022improved,han2020search}. In \cite{shi2020cap}, the authors use bounding boxes instead of rectangular boxes to closely capture an insulator, thereby allowing a neural network to learn the features relevant to the insulator with minimal interference from the background. The work uses dataset of $3700$ images and trained a Faster RCNN model to detect healthy and faulty insulators (with missing disks) and achieved a precision of $\sim90\%$ for the different subsets of their dataset.

Despite the above advances, key challenges still remain unexplored. \textit{First}, articles that address the problem of differentiating between healthy and unhealthy insulators focus only on detecting missing caps. However, insulators can also have discoloured disks as remnants of electrical flashes or broken disks caused by physical damage. The knowledge of these damages is important for the operator to assess the state of health of insulators and prioritise maintenance activities. Detecting patches of discolouration and irregular shapes of disks is a much more difficult task than detecting the presence or absence of an entire cap. The ability of deep neural network models to capture these nuances and detect difficult patterns on insulators is not research in the literature. \textit{Second}, there are several other components such as Stockbridge dampers, bird nests and overgrowing vegetation, the detection of which is important from a monitoring and maintenance perspective of the system operator. Although there have been studies to detect these objects, in particular bird nests \cite{li2020automatic,chen2020bird} and overgrowth of surrounding vegetation \cite{rong2020joint}, the development of a single model capable of detecting a multitude of these objects has not been extensively explored in the literature.

\textit{Finally}, the vast majority of the literature approaches the problem from the perspective of model design, and little effort has been made to augment the richness of the data used to train and evaluate these models. While the development of models for traditional computer vision applications has immensely benefited from benchmark datasets such as Imagenet \cite{deng2009imagenet} and MS-COCO \cite{lin2014microsoft}, inspection of insulators from aerial images faces the challenge of limited data availability in the public domain. Consequently, a large number of insulator inspection models are trained with only one dataset, which is also the target dataset intended for deployment of the model. In such scenarios, the network sees only limited variability in terms of the background and features of objects in the foreground (e.g., different colours of insulator disks, different number and density of disks in the insulator, different types of discs/insulators, different positions of the insulators, etc.). This can result in the network's features not being robust to small changes in the input. The development of a rich dataset that can address this issue is not studied in the literature.

We address the above shortcomings of the literature in this article and develop a multi-object detection model that can learn difficult patterns and identify multiple assets from aerial images. We also prepare a sizable reference dataset collected from multiple open-source repositories that can be used to study the performance of insulator inspection models build with only one target dataset. Such a dataset can also be built upon to create a standrard benchmark dataset for training, evaluation and comparison of different models.

\subsection{Contributions}
This article makes the following contributions:
\begin{enumerate}
    \item \textbf{Detection of flashed and broken disks:} We train standard object detection models to detect healthy, flashed, and broken disks in insulators with a target dataset. This is a much more difficult and important task for planning and scheduling than the tasks previously studied in the literature.
    \item \textbf{Preparation of a reference dataset:} We collect and curate a large dataset from multiple sources that have made their respective datasets publicly available. These datasets contain images of transmission towers with multiple objects of interest, such as insulators, Stockbridge dampers and bird nests. Such a database with images collected by different agencies in multiple countries inherently exhibits a high richness in terms of the features of both foreground and background.
    \item \textbf{Supervised pre-training and fine tuning:} We use this dataset to pre-train standard object detection models in a supervised manner for detecting insulators and classifying them as healthy or faulty. We fine-tune this model to the small target dataset with transfer learning and compare its performance with a model trained from scratch only on the target dataset.
    \item \textbf{Multiple asset detection:} We train end-to-end models for detecting insulators, disks, bird nests and dampers from aerial images with and without transfer learning and compare the performance.
    \item \textbf{Sensitivity analysis:} We use different sizes of training dataset for fine-tuning a pre-trained model and study when having a large reference dataset adds value in terms of performance of the trained model.
\end{enumerate}

The rest of the article is organized as follows: Section~\ref{sec:method} presents the dataset collection and curation procedure adopted to enhance the richness of training data. The approach used to detect multiple objects in images for inspection, the experimental setup and detection tasks performed along with sensitivity analyses are also discussed. Section~\ref{sec:results} presents the results of our experiments  followed by a discussion of the challenges and potential solutions. Finally, Section~\ref{sec:conclusions} provides concluding remarks.

\section{Proposed Method}
\label{sec:method}
In this section, we describe the datasets used for pre-training and fine-tuning object detection models. The different detection tasks considered in this work are then discussed, followed by the setup for conducting different experiments for training and sensitivity analyses of the models.

\subsection{Dataset preparation}
The dataset we use to pre-train object detection models, referred to here as the reference dataset is a combination of three open-source datasets of aerial images of insulators. The first dataset is the Insulator Defect Image Dataset (IDID)~\cite{epriDS2021, vkdw-x769-21} that has $1600$ images at varying resolutions. The median resolution of images in this dataset is $4400\times3008$. Most images in this dataset contain only one insulator, with a few images containing two insulators. The dataset has a considerable variation in the colour of insulator disks (brown, black, white and gray) as well as in the background. The dataset provides the ground truth bounding boxes for insulators, healthy disks, broken disks and flashed disks. The second dataset is the bird nest detection dataset (BND)~\cite{li2020deep,li_jin_2020_4015912} that consists of $401$ images of transmission towers with bird nests along with their ground truth bounding boxes. The median resolution of the images in this dataset is $5472\times3078$. All insulators in this dataset are light green in colour, and the background in most of the images contains crops or grasslands with a few images having small houses and buildings. In contrast to IDID, the BND has images captured from a relatively greater distance and has $2-6$ insulators in most images. These images also contain insulators and Stockbridge dampers. However, this dataset does not provide the ground truth bounding boxes for these objects. The third dataset is the Power Line Assets Detection (STN-PLAD) dataset \cite{vieira2021stn} that consists of $133$ images of transmission towers with insulators and Stockbridge dampers. The median resolution of images in this dataset is $4048\times3040$. This dataset has little to no vegetation in the background and consists mostly of dry patches of flatland and mountains. The insulators and dampers are both of consistent (white) colour and shape in all images in the dataset. The images in this dataset contain $2-4$ insulators and around $10$ dampers and no bird nests. The dataset also provides the ground truth bounding boxes for the insulators and dampers. All the three datasets have images taken with good light exposure and weather conditions, and also have a single type of insulator (either ceramic or polymer). However, the combined dataset has a wide variability in the background as well as foreground.

Owing to the different sources and purposes of the datasets, they contain the ground truths for only a subset of all objects of interest in the images. In order to create a homogeneous and complete dataset, we manually created the ground truth bounding boxes for insulators, nests and dampers for BND with the MATLAB ImageLabeler app~\cite{MATLAB:2021}. As a result, the images from IDID are the only ones that have ground truth for the disks. However, since this dataset constitutes roughly $75\%$ of the reference dataset, this does not pose a significant threat as observed later. The characteristics of the reference dataset are summarised in Table~\ref{tab:source_datasets}, segregated according to the source dataset. It can be observed from Table \ref{tab:source_datasets} that the three datasets combined have a considerable number of objects of each class. Specifically, the resulting reference dataset contains $2647$ insulators, $13364$ healthy disks, $2564$ flashed disks, $1180$ broken disks, $322$ bird nests and $2536$ Stockbridge dampers. In addition, the variations in backgrounds, colours of dampers and insulator disks, distance from the objects and orientation of the camera provide a rich dataset for training object detection models.

\begin{table}[]
    \centering
    \small
    \caption{Source-wise properties of images in the reference dataset (resolution presented as width $\times$ height)}
    \begin{tabular}{c|c|c|c}
        \hline
        Property & \multicolumn{3}{c}{Dataset} \\
        \cline{2-4}
        & IDID & BND & STN-PLAD \\
        \hline
        \hline
        \#Images & 1600 & 401 & 133 \\
        \hline
        Median resolution & $4400\times3008$ & $5472\times3078$ & $4048\times3040$ \\
        \hline
        \#Insulator & 1804 & 531 & 312 \\
        \#Healthy Disk & 13364 & 0 & 0 \\
        \#Flashed Disk & 2564 & 0 & 0 \\
        \#Broken Disk & 1180 & 0 & 0 \\
        \#Bird Nest & 0 & 322 & 0\\
        \#Damper & 0 & 1031 & 1505 \\
        \hline
    \end{tabular}
    \label{tab:source_datasets}
\end{table}

The target dataset considered in this work consist of images provided by the Swiss transmission system operator Swissgrid AG. This dataset contains $77$ images. The background and foreground features of this dataset differ from those in the reference dataset. In order to perform transfer learning and evaluate the performance of fine-tuned models, we manually label the insulators, disks and dampers in this dataset with the LabelImg app \cite{labelimg}, resulting in $279$ insulators, $3706$ healthy disks, $64$ flashed disks, $76$ dampers and $23$ bird nests. This dataset did not have any broken disks. In the following, we discuss the approach adopted to detect insulators, classify them as healthy or faulty as well as perform multi-object detection.

\subsection{Model pool selection}
In this work, we consider four types of object detection models that are popular in the computer vision literature. The first model is Faster RCNN, which is one of the earliest deep neural network-based object detection models. It has also been used in the early works for detection of insulators from aerial images in multiple articles \cite{liu2018insulator,lan2018defect,ma2017detection}. It is a two-stage model that makes use of a region proposal network (RPN) to identify potential object locations in the first stage, followed by another network to adjust the proposed locations and make final predictions in the second stage. The Faster RCNN model suffers from the fact that most of the locations proposed by the RPN do not contain an object, which results in a severe class imbalance problem. This challenge is partly addressed with an adjusted loss function that explicitly accounts for the imbalance, resulting in the RetinaNet model. While the first two models are two-stage object detection models, the third model considered in this work is a single-stage model, called the Fully Convolutional One-Stage detection model (FCOS). FCOS is a relatively recent detection model \cite{tian2019fcos} and is also a proposal-free model that was shown to perform better than its contemporaries. In this article, we consider the Faster RCNN, RetinaNet and FCOS with ResNet50 as the backbone.

The last model considered here is the fifth version of You Only Look Once models, i.e., YOLOv5 which comes from a lineage of highly successful one-stage YOLO models. YOLOv5 performs a series of pre-processing and post-processing steps along with multi-scale detection \cite{glenn_jocher_2022_6222936}, and has also been shown to outperform several other models like the Faster RCNN and RetinaNet. These models are very popular and successful in both computer vision and domain-specific applications, and cover a wide spectrum of modelling approaches including two-stage and one-stage models, imbalance addressing loss function, fully convolutional architectures as well as data augmentation extensive models. We do not consider the recent vision transformer based object detection models that have a very different architecture than convolutional models because of limited size of our datasets.

\subsection{Detection tasks}
\label{subsec:detect}
We are predominantly interested in inspection of insulators. This can be achieved by (a) detecting the insulators and classifying them as healthy or faulty, or (b) detecting all disks and insulators in an end-to-end manner, and identifying healthy and faulty insulators depending on the type of disks detected. Thus, we consider two types of detection tasks for insulators as discussed below. In addition, the use of a deep learning model allows one to detect several objects, which we exploit to detect multiple objects of interest. These tasks discussed in detail below.
\begin{enumerate}
    \item \textit{Task 1: Insulator detection --} The first task involves detection of insulators and is the simplest of the three tasks considered in this article. This task is also the one that is most commonly addressed in the literature, often with modifications to standard object detection models. We use this task to evaluate and compare the performance of the four types of object detection models.
    \item \textit{Task 2: Insulator and disk detection --} The problem of detecting whether an insulator is healthy or faulty can be posed as a classification problem. However, our experiments showed that the performance of a classification-based assessment scheme is very poor. This is potentially due to the fact that there can be different types of damages to the insulator such as discolouration, broken disks and corrosion, each of which results in different patterns on the image. Aggregating different types of patterns into a single class (faulty) can result in the network getting confused at the time of training, and for example, mistaking reflections as discolourations, or different backgrounds as broken disks. Therefore, we pose this problem as one of object detection, where the model is trained to detect different types of disks - healthy, flashed and broken. This approach treats different types of damages in a systematic manner and a post-processing aggregates the classes without interfering in the detection of the disks.
    \item \textit{Task 3: Multiple object detection --} In order to further exploit the potential of deep neural network-based models, we perform a final task of detecting multiple objects of interest. We consider detection of insulators, disks (healthy, flashed and broken), Stockbridge dampers and bird nests, resulting in 6 classes of objects. We demonstrate the potential of object detection models to detect multiple objects with these objects. Although this approach can be extended to any number of object classes, we consider only 6 objects and leave the inclusion of other objects such as corona rings of insulators and spacers for future work.
\end{enumerate}

\subsection{Experimental setup and procedure}
In all our experiments, we resize all the images to $1000\times1000\times3$. The datasets are divided into subsets for training, validation and testing of the models. The exact ratio of these splits vary for different experiments and are detailed later. We adopt random horizontal flipping for data augmentation to train the Faster RCNN, RetinaNet and FCOS models. We observed that including more augmentation methods such as colour jitter and random cropping did not result in a noticeable improvement in performance for the three models. The YOLOv5 model is inherently trained with a host of data augmentation techniques and no additional methods of augmentation were included in the experiments. All our experiments are performed with PyTorch. The built-in models for Faster RCNN, RetinaNet and FCOS in PyTorch are used, while the official repository of YOLOv5 is used in the experiments. The total number of parameters of Faster RCNN, RetinaNet, FCOS and YOLOv5 are $41.29$M, $32.16$M, $32.06$M and $46.5$M respectively. The number of trainable parameters for pre-training is fixed, while this number varies for fine-tuning depending on freezing of different modules of the models. All models are trained (pre-training and fine-tuning) with stochastic gradient descent optimizer with default parameters for $300$ epochs. A batch size of $32$ is used for training Faster RCNN, RetinaNet and FCOS, while YOLOv5 is trained with the default setting of 64 images for backpropagation. The training and evaluation is performed on a workstation with NVIDIA RTX A6000 GPU and $128$ GB of RAM. In order to evaluate the performance of the models, the standard COCO metric of mean average precision (mAP) is adopted. 
We report the mAP with a threshold of 0.5, i.e., mAP$_{50}$ for our experiments. For the sake of presentation, we use mAP and mAP$_{50}$ interchangeably in the rest of the article, while always referring to mAP$_{50}$.

With the above setup in place, we first train detection models with only the target dataset. A (train, val, test) split of ($0.7, 0.0, 0.3$) is used for training Faster RCNN (referred to hereinafter as FRCNN), RetinaNet and FCOS models, while ($0.7,0.1,0.2$) is used for training YOLOv5. We then train detection models with the reference dataset with the above split to obtain the pre-trained models. These models are finally fine-tuned with a (train, val, test) split of ($0.3, 0.0, 0.7$) for FRCNN, RetinaNet and FCOS, while ($0.3, 0.1, 0.6$) for YOLOv5 on the target dataset to obtain the fine-tuned models. We adopt this process for all three detection tasks, resulting in a total of $18$ models. For the sake of brevity, we use the term \textit{target-trained model} to refer to a model trained from scratch and differentiate from pre-trained and fine-tuned models in the rest of the article.
Finally, we vary the training subset size in the set $\{0.1, 0.2, 0.3, 0.4, 0.5, 0.7\}$ and record the performance of the fine-tuned models. In the following section, we present the results of these experiments and provide a discussion of the key insights.

\section{Results}
\label{sec:results}
In this section, we present the results and discuss the findings from our experiments for the three tasks identified above. For each task, we compare different models and training approaches, and highlight the combination that delivers the best performance.

\subsection{Task 1: Insulator detection}
\begin{figure*}[h]
\centering
\subfigure[Faster RCNN]{
    \includegraphics[width=0.23\textwidth]{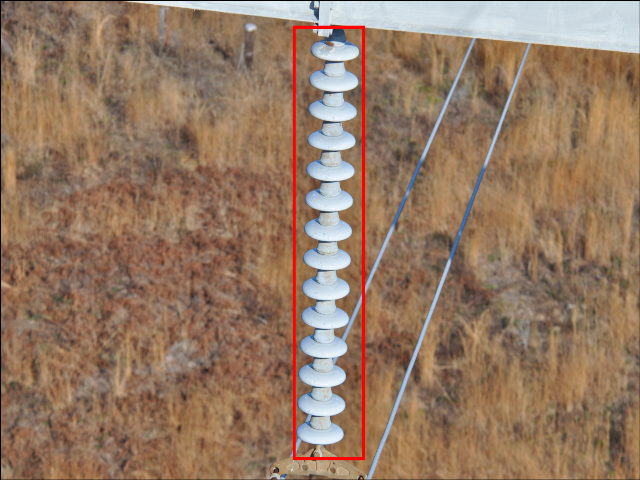}
    \includegraphics[width=0.23\textwidth]{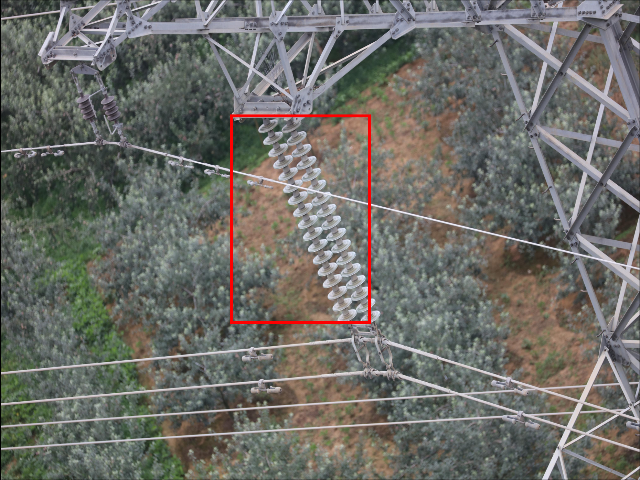}
}
\subfigure[RetinaNet]{
    \includegraphics[width=0.23\textwidth]{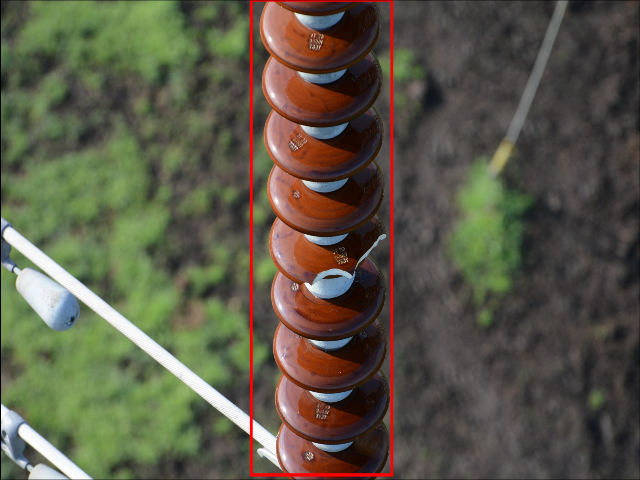}
    \includegraphics[width=0.23\textwidth]{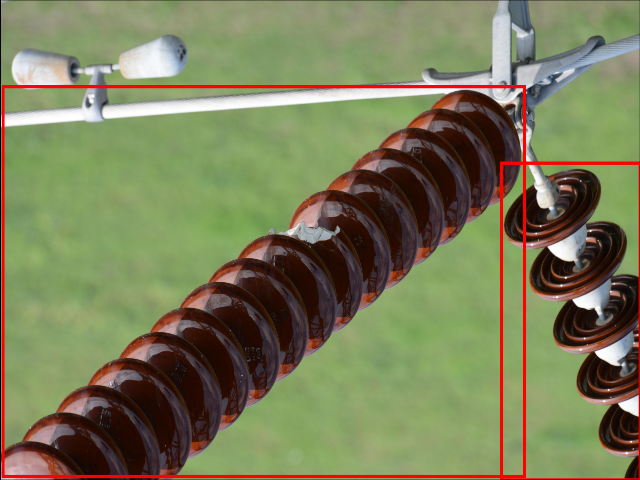}
}
\\
\subfigure[FCOS]{
    \includegraphics[width=0.23\textwidth]{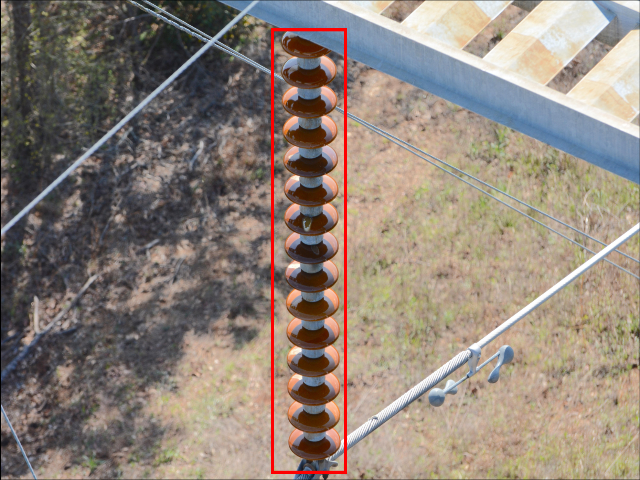}
    \includegraphics[width=0.23\textwidth]{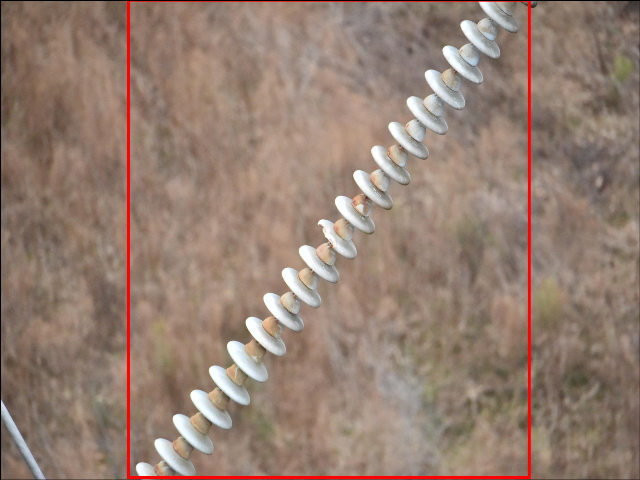}
}
\subfigure[YOLOv5]{
    \includegraphics[width=0.23\textwidth]{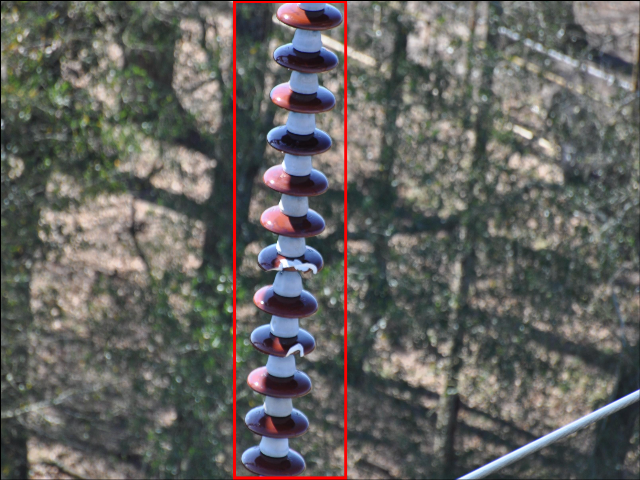}
    \includegraphics[width=0.23\textwidth]{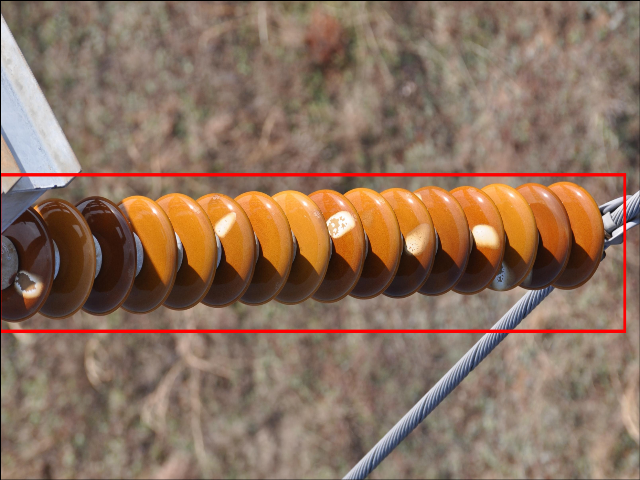}
}
\caption{Predictions of models trained on Task 1 (insulator detection) on test images of reference dataset}
\label{fig:task1}
\end{figure*}

The first and simplest task considered in this article is the detection of insulators. This is a one-class detection problem with fairly large object sizes compared to the other tasks. The mAP$_{50}$ of pre-trained models on the test images of reference dataset are $0.86$, $0.88$, $0.80$ and $0.97$ for FRCNN, RetinaNet, FCOS and YOLOv5 respectively. The YOLOv5 model performs the best, followed by RetinaNet, FRCNN and FCOS. Some illustrative example of predictions with pre-trained models are shown in Fig.~\ref{fig:task1} and the performance of target-trained and fine-tuned models are summarised in Table~\ref{tab:insulator_detection}. The table shows that YOLOv5 outperforms all other models for both types of training. The FRCNN models exhibit the second best performance, followed by FCOS and RetinaNet. It can also be observed that the fine-tuned YOLOv5 model performs much better than others, and is only 14 points short of the model trained from scratch, compared to others that are at least 30 points apart. This is in stark contrast to other models that exhibit a sharp difference in performance between the fine-tuned and target-tuned models.

We observe that YOLOv5 consistently outperforms the other three models for all the tasks and training procedures (training from scratch and fine-tuning). This can be attributed to the data augmentation-heavy pipeline of YOLOv5 which provides a significant advantage to the model in the low data-regime. We therefore focus the ensuing discussion on the performance of YOLOv5 models in the following sections.

\begin{table}[]
    \centering
    \caption{mAP$_{50}$ for Task 1 (insulator detection) on target dataset}
    \vspace{0.1cm}
    \small
    \begin{tabular}{c|c|c}
        \hline
        Model & Trained from scratch & Fine-tuned\\
         \hline
         \hline
         FRCNN & 0.76 & 0.45 \\
         RetinaNet & 0.58 & 0.15 \\
         FCOS & 0.69 & 0.18 \\
         YOLOv5 & 0.87 & 0.73 \\
         \hline
    \end{tabular}
    \label{tab:insulator_detection}
\end{table}


\begin{figure*}[h]
\centering
\subfigure[Faster RCNN]{
    \includegraphics[width=0.23\textwidth]{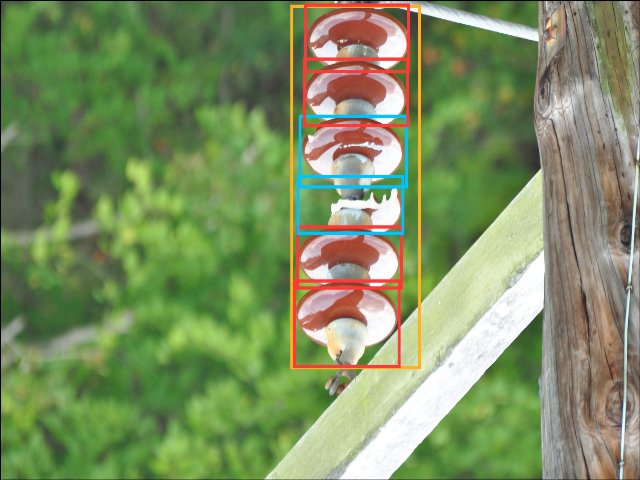}
\includegraphics[width=0.23\textwidth]{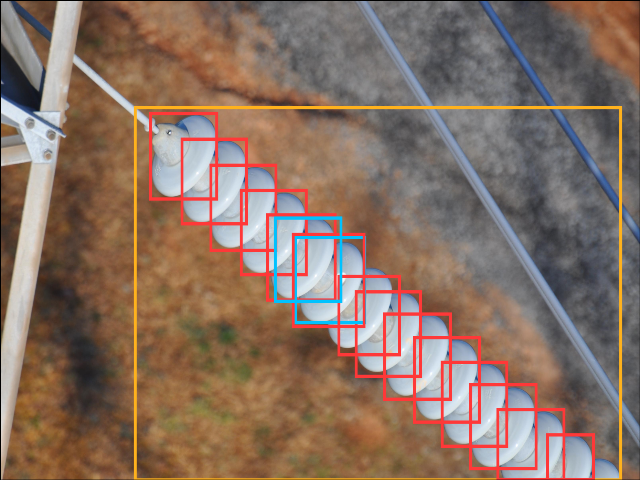}
}
\subfigure[RetinaNet]{
    \includegraphics[width=0.23\textwidth]{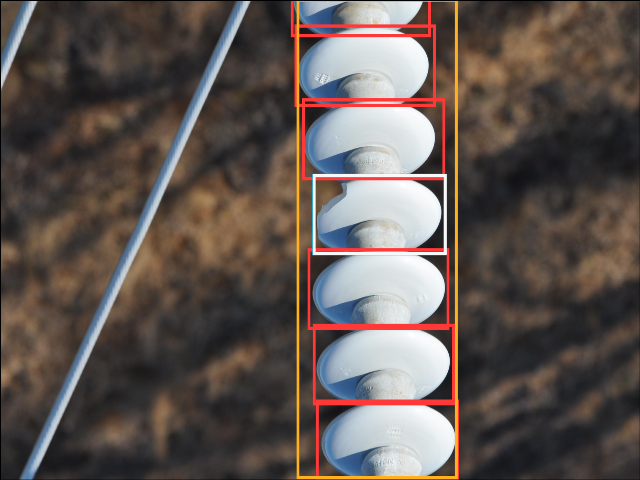}
\includegraphics[width=0.23\textwidth]{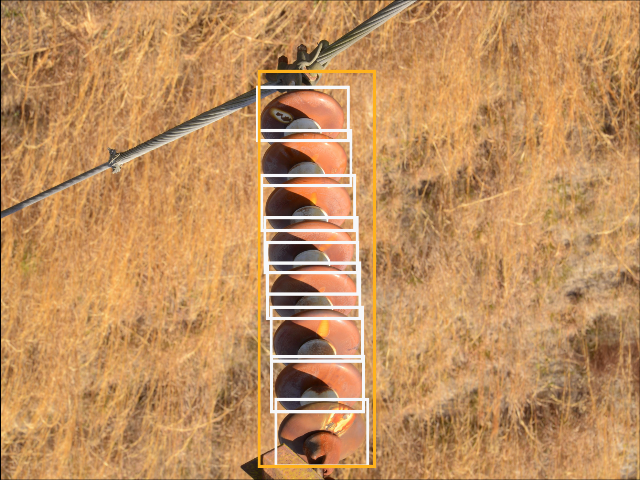}
}
\\
\subfigure[FCOS]{
    \includegraphics[width=0.23\textwidth]{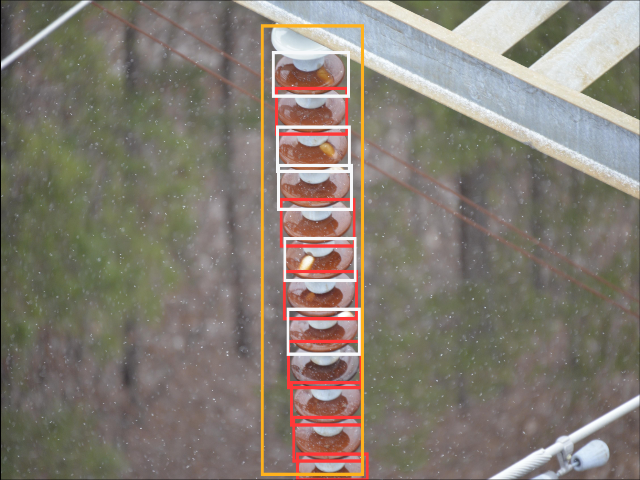}
\includegraphics[width=0.23\textwidth]{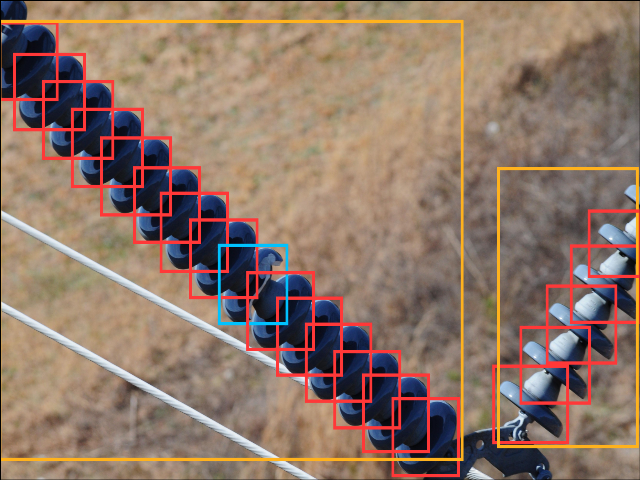}
}
\subfigure[YOLOv5]{
    \includegraphics[width=0.23\textwidth]{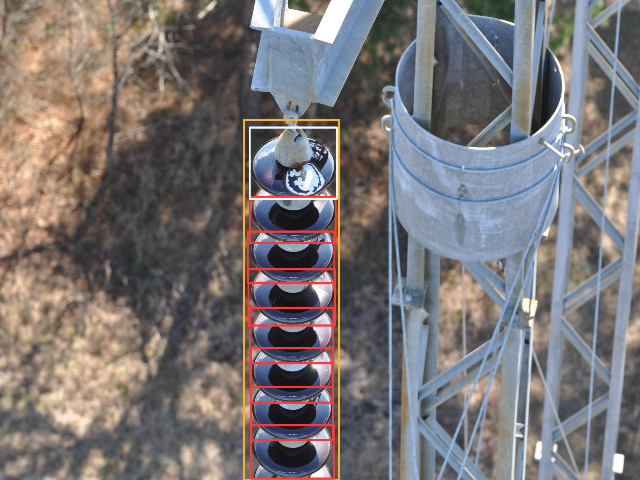}
\includegraphics[width=0.23\textwidth]{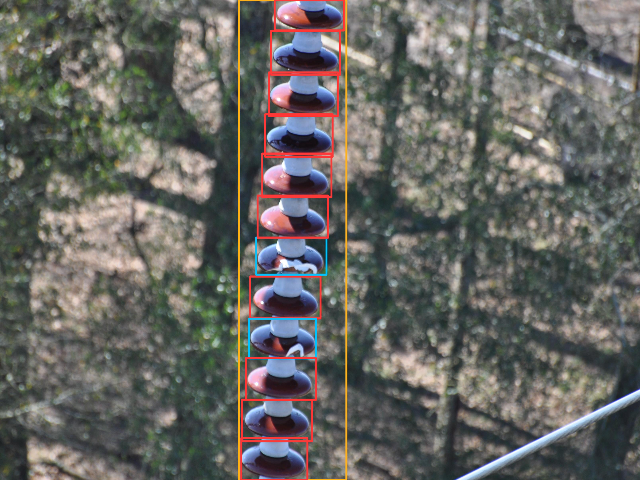}
}
\caption{Predictions of models trained on Task 2 (insulator and disk detection) on test images of reference dataset (orange: insulator, red: healthy disks, blue: broken disks, white: flashed disks)}
\label{fig:task2}
\end{figure*}

\subsection{Task 2: Insulator and disk detection}
The second task involves the detecting healthy and faulty disks in the insulators, which can be used to assess the health of insulators. This is a 4-class detection problem as discussed earlier. In this task, the insulator is of much lager size than the healthy and faulty disks. Since each insulator has approximately $10$ disks, there are always ten times as many disks as insulators. Moreover, the number of healthy disks is also five to six times the number of flashed and broken disks in the reference dataset. Thus, this is a more challenging detection task with varying sizes of the objects and frequencies of their occurrence. The pre-trained YOLOv5 model has an mAP of $0.98$~\footnote{The overall mAP of the model is calculated as an average of the the mAP's of all classes.} on the test images of the reference dataset, while the FRCNN, RetinaNet and FCOS have mAPs of $0.89$, $0.89$ and $0.85$ respectively. These performance scores are comparable to the one-class detection models, with approximately 2-3 points better mAP on the 4-class detection task. This shows that the models are able to capture the differences in patterns of healthy, flashed and broken disks. Some exemplary predictions of the models on reference dataset are illustrated in Fig.~\ref{fig:task2}

The target dataset also exhibits class imbalance across the disks and insulators. The number of healthy disks is about $58$ and $13$ times the number of flashed disks and insulators. This dataset does not have any broken disks. Table~\ref{tab:task2_yolov5} presents the mAP of the pre-trained and fine-tuned YOLOv5 models on the test images of target dataset for the four object types. It can be observed that the models have the worst performance for flashed disks, which are also the least frequent objects in the dataset. The fine-tuned model performs poorer than the target-trained model as also observed for Task 1, while the difference in overall mAP is less (9 points) compared to the one-class detection model.

\begin{table}[]
    \centering
    \caption{mAP$_{50}$ of YOLOv5 models for Task 2 (insulator and disk detection) on target dataset}
    \vspace{0.1cm}
    \small
    \begin{tabular}{c|c|c}
        \hline
        Object & Trained from scratch & Fine-tuned\\
         \hline
         \hline
         Insulator & 0.87 & 0.74 \\
         Disk (H) & 0.76 & 0.72 \\
         Disk (F) & 0.22 & 0.13 \\
         Disk (B) & -- & -- \\
         Overall & 0.62 & 0.53 \\
         \hline
    \end{tabular}
    \label{tab:task2_yolov5}
\end{table}

\subsection{Task 3: Multiple object detection}
The third and final task involves detection of six objects and is the most difficult task. Both the reference and target datasets have severe class imbalance for bird nests in addition to the flashed and broken disks. The Stockbridge dampers are also the smallest objects as well as up to five times less frequent than the healthy disks. The mAP of the pre-trained models on the test images of reference dataset are $0.85$, $0.85$, $0.79$ and $0.85$ for the FRCNN, RetinaNet, FCOS and YOLOv5 models. These scores are also comparable to the one-class and four-class models, allowing for detection of multiple assets from the aerial images.

\begin{figure*}[]
    \centering
    \includegraphics[width=0.47\textwidth]{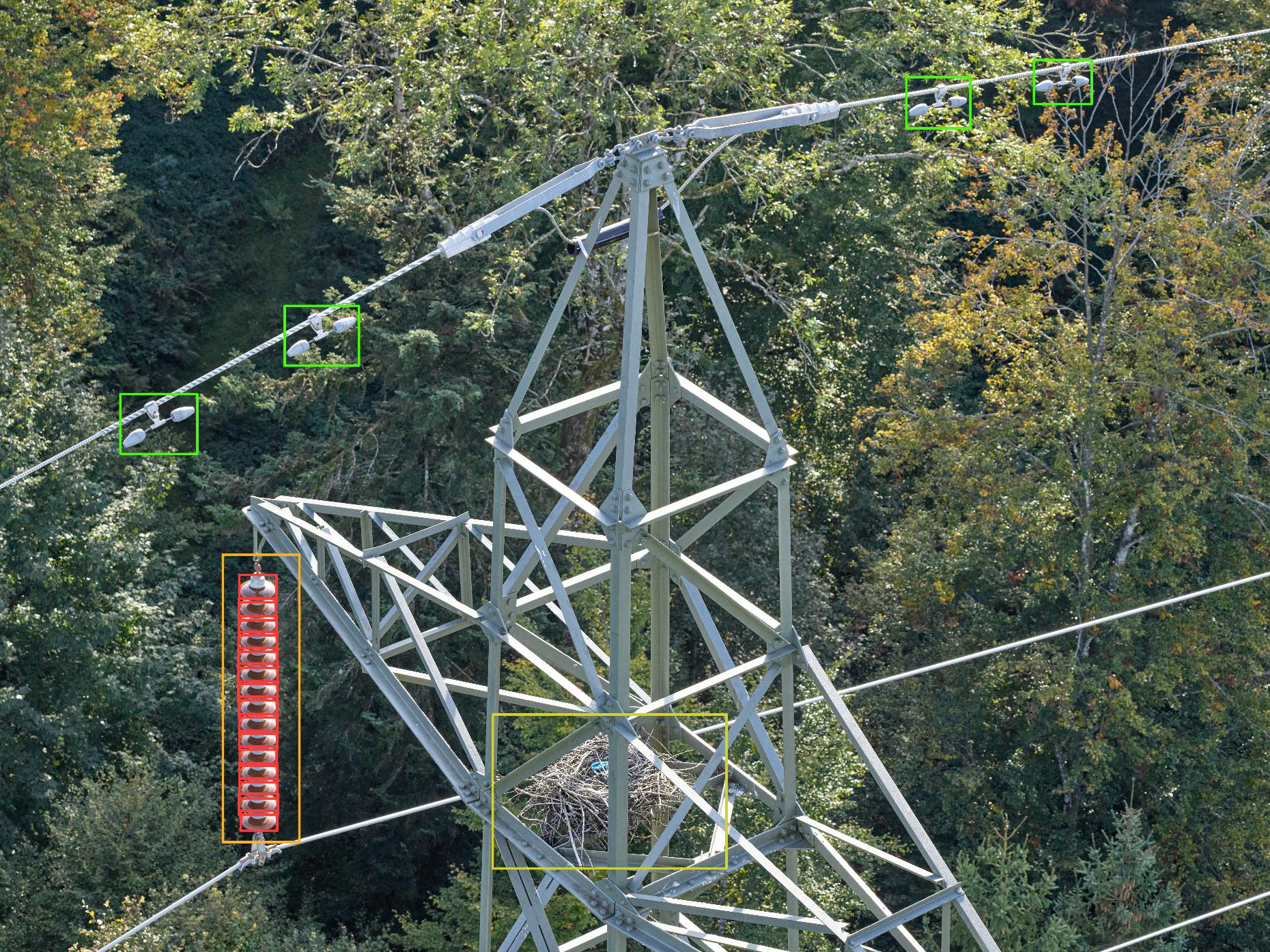}
    \includegraphics[width=0.47\textwidth]{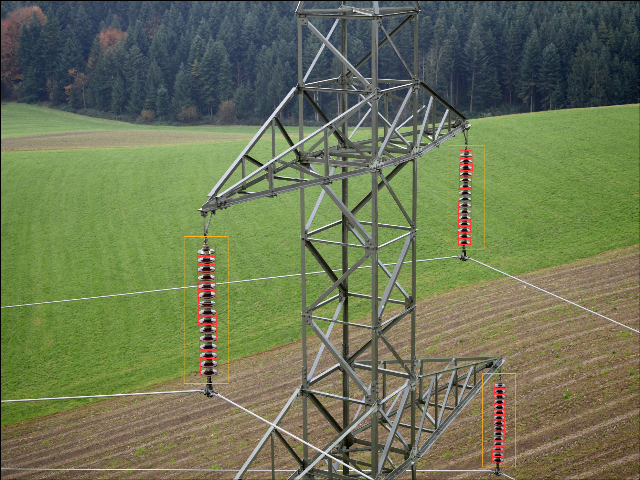}
    \caption{Predictions of YOLOv5 model trained from scratch on Task 3 (multiple object detection) on test images of target dataset (orange: insulator, red: healthy disks, green: Stockbridge dampers, yellow: bird nest)}
    \label{fig:task3_yolov5}
\end{figure*}

Table~\ref{tab:task3_yolov5} summarises the mAP of the target trained and fine-tuned YOLOv5 models. It can be observed that the the difference in performance between the two models is much more severe than the one-class and four-class models. Specifically, the overall mAP of the fine-tuned models is 18 points lower than that of the target-trained model, the insulator and bird nest being the biggest contributors to this difference. This can potentially be due to the nature of these objects. Specifically, the bird nest is always occluded by structures of the tower and might be difficult to detect - even by humans. The insulator, on the other other hand is the largest object, which stands out from all the other objects that are relatively small. This difference in size compared to all other classes in addition to the class imbalance might potentially lead the model to emphasise more on predicting objects at a smaller scale than at a larger scale. This can potentially be addressed by appropriately weighing the contributions of the objects to the loss function. Figure \ref{fig:task3_yolov5} shows the predictions of the model on two sample test images of target dataset. The bounding boxes predicted by the model for different objects are shown in different colours. It can be seen that the model is able to detect all the healthy disks, dampers, insulators and a bird nest in the images. However, in the right image, the model misclassifies a flashed disk as healthy, which explains its poor performance in detecting flashed disks on target dataset (see Table~\ref{tab:task3_yolov5}).
\begin{table}[]
    \centering
    \caption{mAP$_{50}$ of YOLOv5 models for Task 3 (multiple object detection) on target dataset}
    \vspace{0.1cm}
    \small
    \begin{tabular}{c|c|c}
        \hline
        Object & Trained from scratch & Fine-tuned\\
         \hline
         \hline
         Insulator & 0.90 & 0.67 \\
         Disk (H) & 0.77 & 0.70 \\
         Disk (F) & 0.18 & 0.14 \\
         Disk (B) & -- & -- \\
         Nest & 0.87 & 0.48 \\
         Damper & 0.71 & 0.50 \\
         Overall & 0.68 & 0.50 \\
         \hline
    \end{tabular}
    \label{tab:task3_yolov5}
\end{table}

\subsection{Sensitivity analysis}
In the previous sections, we presented the performance of the fine-tuned models that used $30\%$ of target dataset for training. On the other hand, the model trained from scratch uses $70\%$ of the target dataset. It is common practice to use a smaller portion of the target dataset for fine tuning, and is one of the key advantages of using a pre-trained model. However, we observe that in this application, many of the fine-tuned models perform significantly worse than the target-trained models, as seen in Table~\ref{tab:insulator_detection}, thus discouraging the use of pre-training and fine-tuning. We perform a sensitivity analysis to examine whether the amount of training data has any impact on the performance of the fine-tuned models. We use $10\%$, $30\%$ and $70\%$ of the target dataset for fine-tuning and record the performance of the models on the remaining images. Table~\ref{tab:sensitivity_yolo} lists the object-wise mAP of these fine tuned YOLOv5 models. Table~\ref{tab:sensitivity_yolo} shows that even with $10\%$ of the data, i.e., $8$ images used for fine-tuning, the model has mAPs of $0.63$ and $0.7$ for the insulator and healthy disk respectively. This can be attributed to the large size of insulators and high frequency of healthy disks. The detection of rare classes is unsurprisingly poor in the beginning, but improves significantly with addition of more images. The last model uses the same number of images for fine-tuning as the model trained from scratch. These two models differ only in the initial parameters before training, i.e., while the former starts with parameters learnt from the reference dataset, the latter starts with a random initialisation. A comparison of the two models (Table~\ref{tab:task3_yolov5} and Table~\ref{tab:sensitivity_yolo}) reveals that the fine-tuned model has poor performance compared to the target-trained model even when trained on the same images. We observe the opposite behaviour for the Faster RCNN, RetinaNet and FCOS models across all tasks, with approximately 5 points improvement in mAP compared to the target-trained model. This difference in behaviour can potentially be due to the different data processing pipeline of YOLOv5.
\begin{table}[]
    \centering
    \caption{mAP$_{50}$ of target-trained and fine-tuned YOLOv5 models trained with different fractions of target dataset for Task 3 (multiple object detection); (TT: target-trained, FT: fine-tuned).}
    \vspace{0.1cm}
    \small
    \begin{tabular}{c|c|c|c|c|c|c}
        \hline
         \multirow{2}{*}{Object}& \multicolumn{2}{c|}{0.1} & \multicolumn{2}{c|}{0.3} & \multicolumn{2}{c}{0.7} \\
         \cline{2-7}
         & TT & FT & TT & FT & TT & FT \\
         \hline
         \hline
         Insulator & 0.74 & 0.63 & 0.70 & 0.67  & 0.90 & 0.82\\
         Disk (H) & 0.62 & 0.70 & 0.77 & 0.70 & 0.77  & 0.60\\
         Disk (F) & 0.04 & 0.08 & 0.07 & 0.14 & 0.18 & 0.19\\
         Disk (B) & -- & -- & -- & -- & -- & --\\
         Nest & 0.11 & 0.29 & 0.47 & 0.48 & 0.87 & 0.66\\
         Damper & 0.06 & 0.07 & 0.83 & 0.50 & 0.71 & 0.58\\
         Overall & 0.31 & 0.36 & 0.57 & 0.50 & 0.68 & 0.57\\
         \hline
    \end{tabular}
    \label{tab:sensitivity_yolo}
\end{table}

\begin{figure}
    \centering
    \includegraphics[width=0.46\textwidth]{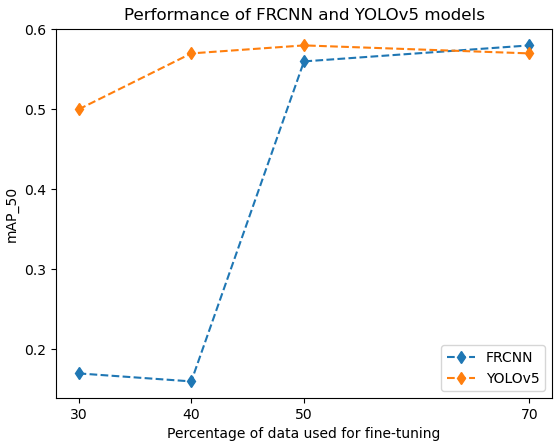}
    \caption{Test performance of FRCNN and YOLOv5 models (Task 3) fine-tuned with different fractions of target dataset}
    \label{fig:sensitivity}
\end{figure}

Figure~\ref{fig:sensitivity} compares the performance of fine-tuned FRCNN and YOLOv5 models for Task 3 with different fractions of target dataset. We observe that the YOLOv5 model performs much better in the extremely low-data part of the plot. This can potentially be attributed to the data augmentation-heavy pipeline of YOLOv5, which adds significant value with $40\%$ and smaller fractions of datasets used for training. On the other hand, the FRCNN model has very poor performance in the beginning and performs comparably with the YOLOv5 model when more data is available.

\subsection{Discussion}
In this section we discuss a few insights from the results and outline key aspects that can be addressed in the future. We performed pre-training with a feature-rich reference dataset with four different object detection models for three detection tasks and found that all models performed well on detection tasks with varying levels of difficulty. This suggests that object detection models can be used to detect difficult patterns such as flashed and broken disks compared to those previously studied in the literature. Detecting these incipient faults can prove helpful to system operators for planning maintenance activities. We also found that YOLOv5 performed better in most tasks, with the Faster RCNN model performing comparably for multiple asset detection.

We used the pre-trained model to fine-tune to a small target dataset and examined if transfer learning adds value compared to training from scratch in a low data-regime. The models trained from scratch performed better than those obtained with fine-tuning. On the other hand, Both types of training resulted in poor performance for under-represented classes in terms of both frequency and size of the objects. While the imbalance in frequency is addressed by the focal loss in RetinaNet, the multi-scale detection pipeline of YOLOv5 allows detecting objects of different sizes. Despite these features, these models performed relatively poorly on the under-represented objects.

We performed a sensitivity analysis and found that the impact of amount of data depends on the type of the model. Specifically, using as much data as training from scratch to fine tune a pre-trained model results in an improvement in performance by about 5 points for Faster RCNN, RetinaNet and FCOS models. However, the YOLOv5 models exhibit the opposite behaviour with the target-trained model always performing better than the fine-tuned models. This is of particular importance and suitable for extremely low data applications, where the target dataset has very few samples. We also observed that the augmentation-heavy YOLOv5 model outperforms FRCNN for small fractions of dataset, while the latter catches up with availability of more images. In addition, these results also validate the importance of the reference dataset used for obtaining the pre-trained models.


The above results can further be improved in several ways. First, the reference dataset consists of three different sources, and therefore has incomplete labels. The IDID has labels for different types of disks, while BND and STN-PLAD do not have labels for any disks. As a result, the pre-trained model learns to identify disks for the former dataset, and at the same time not to predict the disks of the latter two datasets. This can be addressed by completing the labels of all the images in the reference dataset. Second, the target dataset does not have any broken disks, and as a result the effect of different training methods on the mAP of detecting them has not been studied. This can be pursued in the future. Third, no architectural modifications have been made to the object detection models. This can be explored further to identify models that better capture the objects at different scales occurring in different frequencies. Finally, the use of state-of-the-art generative models for augmenting the target dataset in a low data-regime, and incorporating these synthetic images in the training and fine-tuning can be explored.

\section{Conclusion}
\label{sec:conclusions}
In this article, we present the results of object detection-based inspection of insulators from aerial images. We identify flashed and broken disks as important types of incipient faults in insulators, that are difficult to detect compared to the faults considered in the literature (missing caps). We collect and curate a reference dataset from three different repositories that exhibit rich features of the objects of interest, as well as variations in the background. We train four object detection models that are able to detect these incipient faults with very good accuracy.

We perform three inspection tasks with four types of object detection models trained from scratch on a small target dataset, as well as pre-trained on the reference dataset followed by fine-tuning on the target dataset. Our experiments reveal that the YOLOv5 model outperforms the Faster RCNN, RetinaNet and FCOS models. Sensitivity analysis of the amount of data used for fine-tuning suggests that YOLOv5 derives its better performance from a data augmentation-heavy pipeline, which is important in the low data-regime. We also identified that the impact of dataset size for fine-tuning depends on the type of object detection model used. The findings of the article highlight the value of using pre-training in low data-regimes, identify frequency and size imbalance as the major causes of poor performance, and point out potential directions for future research to improve the state-of-the-art in object detection-based insulator inspection in the low data-regime.

\section*{Acknowledgements}
\label{Acknowledgements}
This work was supported by the Swiss Federal Office of Energy: “IMAGE - Intelligent Maintenance of Transmission Grid Assets” (Project Nr. SI/502073-01). The authors would additionally like to thank the student assistants, Ms. Manon Prairie and Mr. Tyler Anderson for generating the ground truth labels for the target dataset.
 \bibliographystyle{elsarticle-num} 
 \bibliography{cas-refs}





\end{document}